\documentclass[runningheads]{llncs}
\usepackage[T1]{fontenc}
\usepackage{graphicx}
\usepackage{hyperref}
\usepackage{color}
\usepackage[strings]{underscore}
\usepackage{booktabs}
\usepackage{float}
\usepackage{amsmath}
\usepackage{subcaption}

\begin{document}
\title{ProcessTBench: An LLM Plan Generation Dataset for Process Mining}
\titlerunning{ProcessTBench: An LLM Plan Generation Dataset for Process Mining}
\author{Andrei Cosmin Redis\inst{1} \and
Mohammadreza Fani Sani\inst{2} \and
Bahram Zarrin\inst{2} \and
Andrea Burattin\inst{1}}
\authorrunning{AC Redis et al.}
\institute{Technical University of Denmark\\ \email{andreiredis@gmail.com} \\ \email{andbur@dtu.dk}\\ \and
Microsoft Development Center Copenhagen\\ 
\email{mfanisani@microsoft.com}\\
\email{bahram.zarrin@microsoft.com}}
\maketitle   

\begin{abstract}
Large Language Models (LLMs) have shown significant promise in plan generation. Yet, existing datasets often lack the complexity needed for advanced tool use scenarios — such as handling paraphrased query statements, supporting multiple languages, and managing actions that can be done in parallel. These scenarios are crucial for evaluating the evolving capabilities of LLMs in real-world applications. Moreover, current datasets don’t enable the study of LLMs from a process perspective, particularly in scenarios where understanding typical behaviors and challenges in executing the same process under different conditions or formulations is crucial. To address these gaps, we present the ProcessTBench synthetic dataset, an extension of the TaskBench dataset specifically designed to evaluate LLMs within a process mining framework.

\end{abstract}

\vspace{-1 cm}
\section{Introduction}

The rapid advancements in Large Language Models (LLMs) have generated significant interest in their potential across various domains, particularly in their tool use and plan generation capabilities. These capabilities are increasingly critical as LLMs are envisioned to provide natural language interfaces for complex process automation. Despite their promise, the emergent properties of LLMs remain not fully understood, making empirical studies essential for advancing LLM plan generation, as evidenced by the growing number of benchmarks aimed at testing various LLM use cases~\cite{chang_survey_2024}.

Despite the promise of LLMs, plan generation remains in its early stages, often struggling with the reliability needed for complex tasks. This limitation highlights the importance of evaluating LLM behavior on sophisticated tasks to better equip them for executing advanced processes. Existing benchmarks, such as ToolBench~\cite{xu_tool_2023}, have made valuable progress in this area, but they frequently overlook the intricacies of LLM-generated plans, including sequence length, parallelism, and the handling of paraphrased queries. The absence of datasets featuring multiple rephrasings of the same query limits our ability to assess LLMs’ robustness and adaptability. Additionally, it constrains our understanding of LLMs from a process perspective, particularly in scenarios requiring consistent execution of the same process under varying conditions or formulations. This is especially vital where LLM agents must adhere to strict, predefined processes, ensuring conformance to established guidelines and procedures.
Furthermore, in certain applications, there may not be any reference process models available for the plans provided. Consequently, devising a method to identify and depict past or potential future behaviors could enhance the clarity of actions for decision-makers.

Building on the foundation of TaskBench~\cite{shen_taskbench_2024}, our work introduces the ProcessTBench synthetic dataset \footnote{https://github.com/microsoft/ProcessTBench} to address existing gaps by providing a more challenging environment for evaluating LLMs in plan generation from a process mining perspective. While TaskBench offers a challenging query dataset with ground truth solutions, ProcessTBench extends this by incorporating multi-lingual query paraphrases, a plan generation framework tailored for process mining, and a comprehensive synthetic planning dataset. The synthetic dataset includes queries that demand not only complex sequences and parallel actions but also involve paraphrases of the problem query. Additionally, it provides the synthetic planning dataset in a sequential, text-based format, enabling a process mining analysis of LLM behavior, particularly in how these models handle tool use across various scenarios and adapt to different query formulations.

With this synthetic dataset, we aim to support research into analyzing LLM planners’ behavior, identifying common pitfalls, and exploring opportunities for improvement through process mining techniques. Ultimately, we hope to foster a more nuanced understanding of LLM plan generation and provide the community with a resource for developing and testing sophisticated LLM frameworks in complex and dynamic environments.

\section{Generating ProcessTBench} 
The data generation pipeline~\ref{fig-generation} consists of the following components:

\begin{figure}[hbt!]
    \centering
    \vspace{-1.3cm}
    \includegraphics[width=0.9\linewidth]{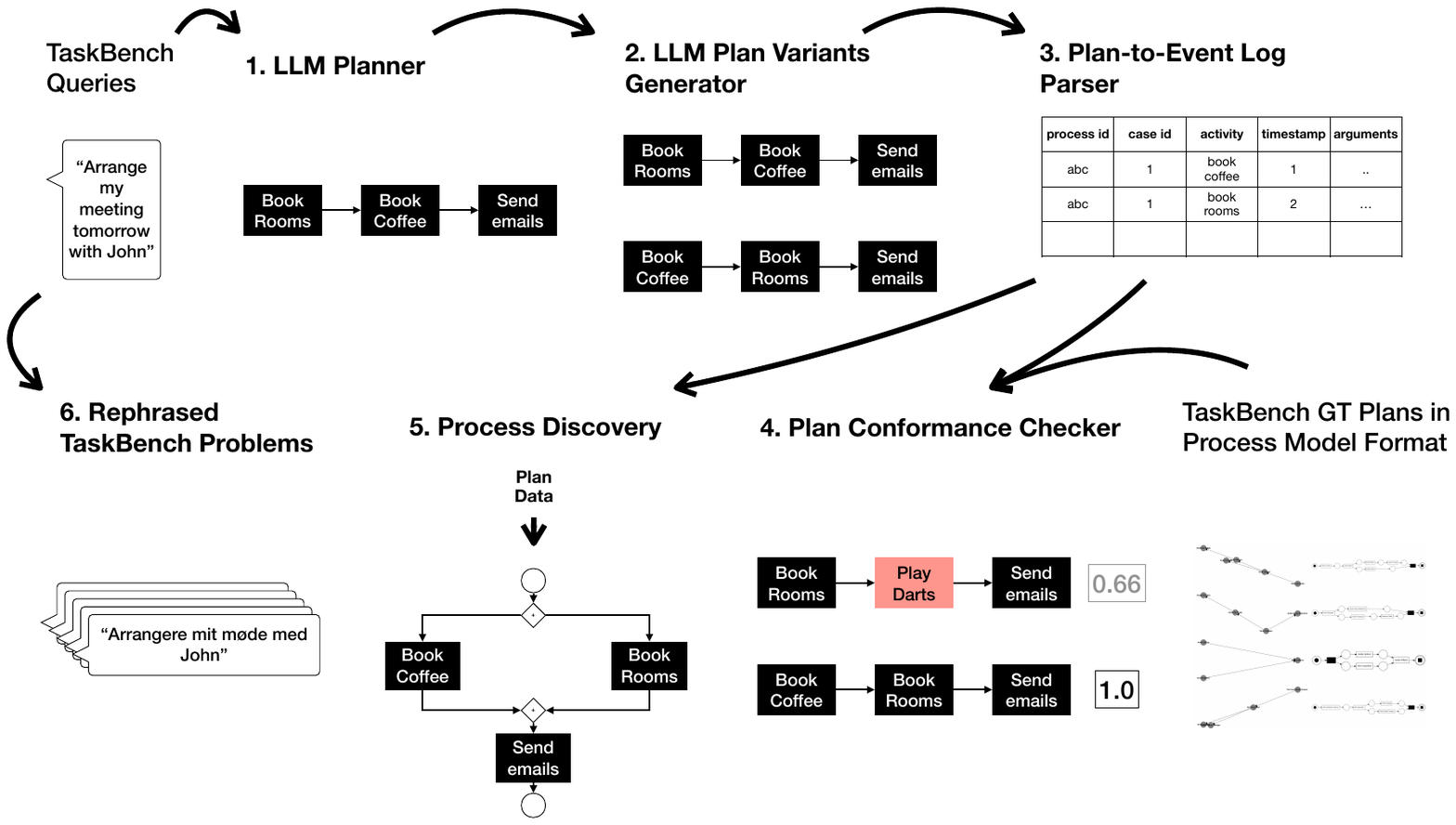}
    \vspace{-1.3cm}
    \caption{The data generation pipeline of ProcessTBench. The arrows symbolize that the output of one step is input to the next.}
    \label{fig-generation}
    \vspace{-0.3cm}
\end{figure}

\begin{enumerate}
\item [*] \textbf{TaskBench Queries and Ground Truth Plans in Process Model Format}. We selected the most challenging subset of the TaskBench dataset (TaskBench Multimedia), comprised of 565 queries and their respective ground truth plans as directed acyclic graphs.
\item \textbf{LLM Planner}. Given a query and a set of available tools, it generates a sequence of tool invocations to resolve the query~\ref{tab-eventlog}. It is slightly modified from planners like ReAct~\cite{yao_react_2022}, in that with one inference, it generates all the tool invocations necessary to solve the query instead of only the next step. 
\begin{enumerate}
    \item \textit{Input}: query text and available tools in text format. Only the tools required to solve the query were given.
    \item \textit{Output}: Plan that solves the query with the given tools~\ref{tab-eventlog}
\end{enumerate}
\item \textbf{LLM Plan Variants Generator}. Given the query, available tools, and a plan, generate alternative plans to solve the query. This is done to create more cases for the event log. The prompt is provided in the code repository of ProcessTBench.
\begin{enumerate}
    \item \textit{Input}: Input and Output of step 1
    \item \textit{Output}: Alternative plans that solve the query
\end{enumerate}
\item \textbf{Event Log Parser}. Parses the plans created into an event log usable for process mining.
\begin{enumerate}
    \item \textit{Input}: Output of steps 1 and 2
    \item \textit{Output}: Event log, as described in~\cite{van_der_aalst_process_2016}.
\end{enumerate}
\item \textbf{Plan Conformance Checker}. This downstream task is an example use case of our synthetic dataset. Using conformance checking, as in Chapter 8 of~\cite{van_der_aalst_process_2016},  this component verifies the traces generated by the planner against the corresponding ground truth process model provided in TaskBench. The metrics used for validation are replay and alignment fitness, as seen in the Chapter 8 of~\cite{van_der_aalst_process_2016}. Furthermore, the ground truth plans in TaskBench, represented as directed acyclic graphs (DAGs), were converted to Petri nets, the standard plan (process) format for conformance checking. The conversion algorithm is provided in the code repository of ProcessTBench.
\begin{enumerate}
    \item \textit{Input}: Output of step 3
    \item \textit{Output}: Event log with alignment and replay fitness for each case (plan) generated.
\end{enumerate}
\item \textbf{Process Discovery}. This downstream task is an example use case of our synthetic dataset. Using the inductive miner, this component generates process models as Petri nets from the event log, as in Chapter 6 of~\cite{van_der_aalst_process_2016}. 
\begin{enumerate}
    \item \textit{Input}: Output of step 3
    \item \textit{Output}: Process models as petri nets
\end{enumerate}
\end{enumerate}

The data was generated between Sep. 2023 and Feb. 2024. The LLM model used throughout is gpt-4-0613. More experimental details are available in the code repository provided.
\vspace{-0.3cm}
\section{Description of ProcessTBench}
\vspace{-0.1cm}

The ProcessTBench synthetic dataset builds upon TaskBench~\cite{shen_taskbench_2024}, focusing on task complexity, tool usage, and process characteristics. The ProcessTBench synthetic dataset includes 532 base queries drawn from the most challenging subset of TaskBench, each paraphrased 5 to 6 times, with an average of 4.08 solution plans per query. These plans involve action sequences utilizing a subset of 40 unique tools. Corresponding to the queries, ProcessTBench additionally includes the respective ground truth plans in Petri net format.
\vspace{-0.2cm}
\subsection{Queries}
\begin{figure}[hbt!]
    \centering
    \begin{subfigure}[t]{0.45\linewidth}
        \centering
        \includegraphics[width=\linewidth]{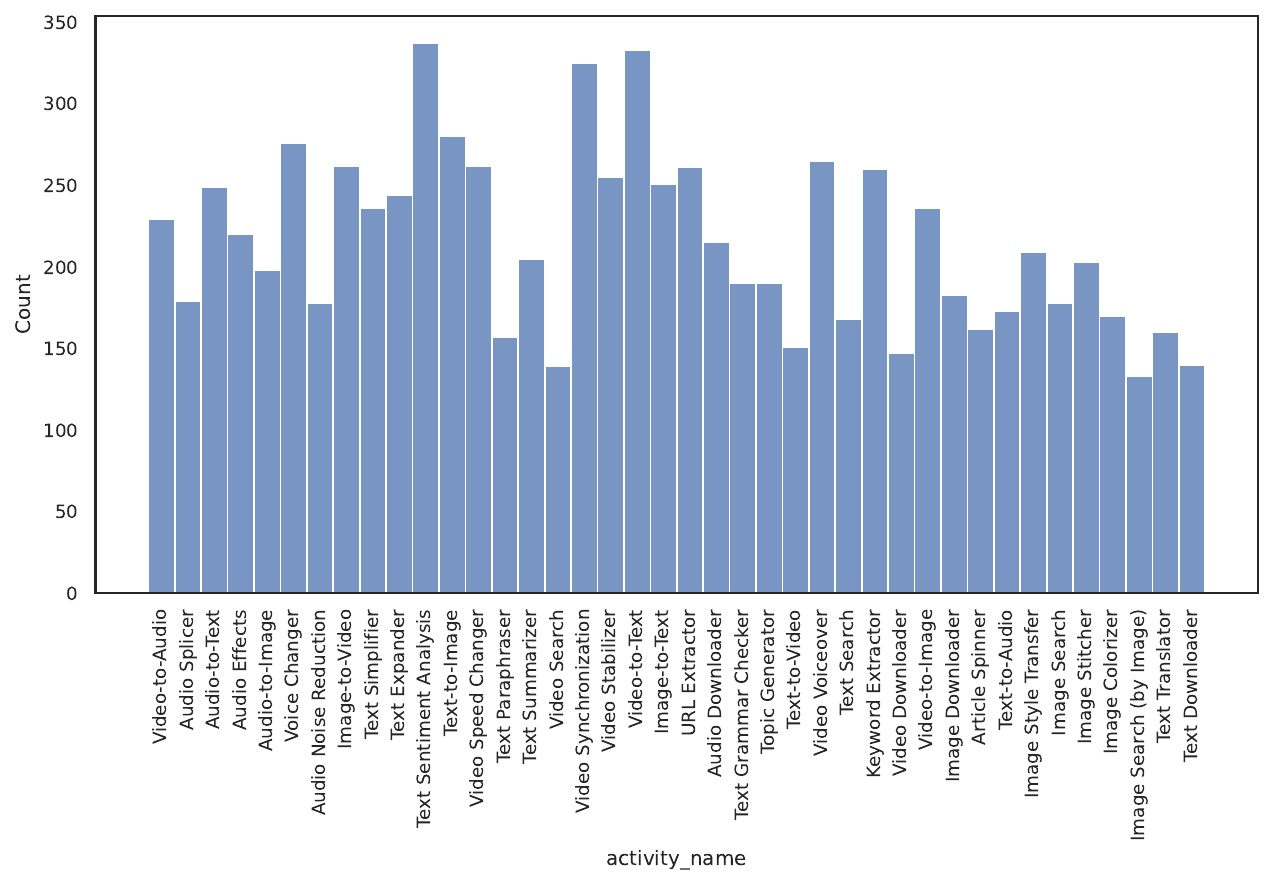}
        \caption{Distribution of actions in the ProcessTBench synthetic dataset.}
        \label{fig-distribution}
    \end{subfigure}
    \hfill
    \begin{subfigure}[t]{0.45\linewidth}
        \centering
        \includegraphics[width=\linewidth]{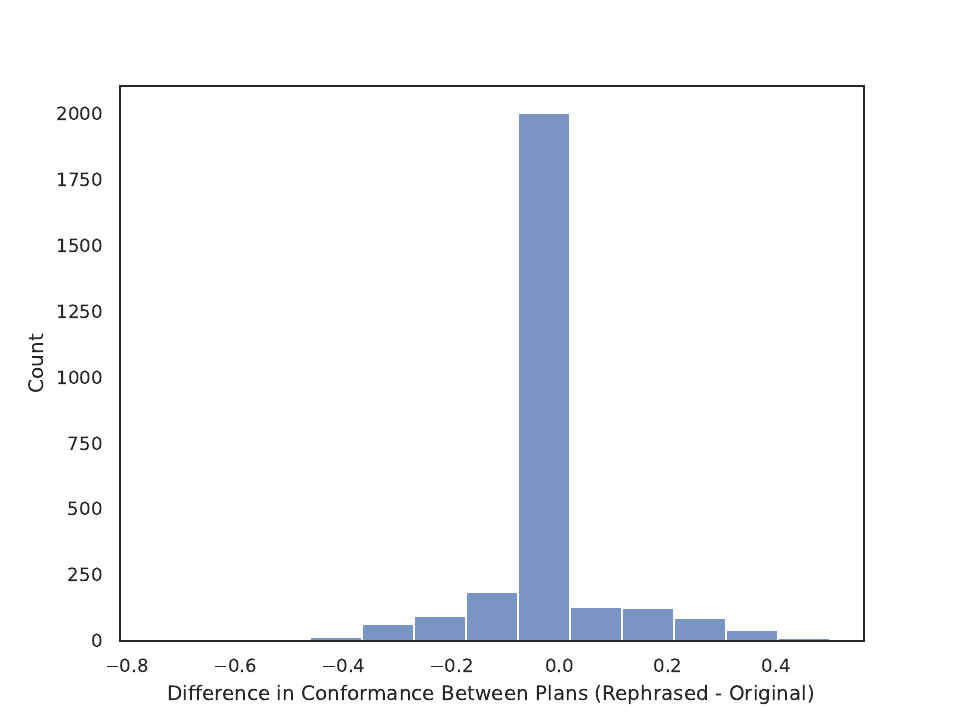}
        \caption{The difference in alignments between plans from queries and their paraphrases}
        \label{fig-plan-dist}
    \end{subfigure}%
    \caption{Key quality features of the queries in ProcessTBench.}
\end{figure}

\vspace{-0.1cm}
To validate the quality of the base queries selected from TaskBench, we present a balanced distribution of actions required to solve these queries, as shown in Figure~\ref{fig-distribution}. This distribution suggests an even representation of task types, ensuring comprehensive coverage across various action categories.

Additionally, we evaluated the quality of the paraphrased queries by using an LLM plan generator to create plans for both the original TaskBench queries and their paraphrased counterparts. We then applied conformance checking, specifically alignment fitness (as described in Chapter 8 of~\cite{van_der_aalst_process_2016}), to compare the generated plans. Figure~\ref{fig-plan-dist} illustrates the alignment differences between the original and paraphrased queries, showing that the paraphrased queries generally maintain equivalent alignment quality. Specifically, the comparison revealed 1,965 instances of equivalence, 397 instances where the paraphrased queries performed worse, and 389 instances where they performed better. The mean alignment difference was 0.00 with a standard deviation of 0.11. A Wilcoxon signed-rank test yielded a p-value of 0.56, suggesting that there is no significant difference in the quality between the original and paraphrased queries.
 
\vspace{-0.3cm}
\subsection{Ground Truth and Generated Plans}
\vspace{-0.1cm}
Table~\ref{tab-summary} provides an overview of the process-related characteristics of both the ground truth and generated plans within ProcessTBench. Each query in ProcessTBench is associated with a ground truth plan in Petri net format and 5-6 LLM-generated plans derived using a custom prompt. The complexity of these ground truth Petri nets is quantified using Cardoso and Cyclomatic complexity metrics~\cite{lassen_complexity_2009}, enabling comparison with other process mining datasets. The degree of concurrency, defined as the ratio between the longest and shortest paths in a Petri net, measures the level of parallelism (a ratio of 1.00 indicates no parallelism, while values greater than 1 signify increasing levels of parallel behavior). Additionally, the mean number of plan variants reflects the diversity of alternative solutions generated for each query. One sample from the synthetically generated plan dataset is shown in Table~\ref{tab-eventlog}.

\begin{table}[hbt!]
    \vspace{-0.6 cm}
    \centering
    \scriptsize
    \begin{tabular}{lrrrrr}
    \toprule
     & mean & std & min & 50\% & max \\
    \midrule
    Cases (Plans) / Process& 4.08 & 1.27 & 2 & 4 & 11 \\
    Case Variants / Process& 2.68 & 0.97 & 1 & 3 & 5 \\
    Case (Plan) Length & 3.79 & 0.89 & 2& 4& 8\\
    Cardoso Complexity of GT Process& 6.92 & 1.32 & 2& 7& 14\\
    Cardoso Complexity of GT Process & 8.86 & 5.40 & 2& 7& 74\\
    Degree of Concurrency of GT Process & 1.43 & 1.00 & 1.00 & 1.33 &3.00 \\
    \bottomrule
    \end{tabular}
\caption{Summary of ProcessTBench synthetic dataset features from a process model perspective, including statistics on the number of cases and variants per process, plan length, Cardoso complexity, and degree of concurrency. } 
\label{tab-summary}
\end{table}

\begin{table}[hbt!]
\vspace{-1.6 cm}
    \centering
    \resizebox{\textwidth}{!}{
    \begin{tabular}{lrlll}
    \toprule
    process id &  case id &  activity name &  timestamp &  arguments \\
    \midrule
    36690562& 1 & Video-to-Audio& 02/15/2024 @ 14:55& ['example.mp4']\\
    36690562& 1 & Audio Splicer& 02/15/2024 @ 15:00& ['<Video-to-Audio>', 'example.wav']\\
    36690562& 1 & Audio-to-Text& 02/15/2024 @ 15:01& ['<Audio Splicer>']\\
    36690562& 1 & Audio Effects& 02/15/2024 @ 15:02& ['<Audio Splicer>', 'add reverb']\\
    36690562& 2 & Video-to-Audio& 02/15/2024 @ 16:55& ['example.mp4']\\
    36690562& 2 & Audio Splicer& 02/15/2024 @ 17:00& ['<Video-to-Audio>', 'example.wav']\\
    36690562& 2 & Audio Effects& 02/15/2024 @ 17:01& ['<Audio Splicer>', 'add reverb']\\
    36690562& 2 & Audio-to-Text& 02/15/2024 @ 17:02& ['<Audio Splicer>']\\
    \bottomrule
    \end{tabular}}
    \caption{A sample of the event log from the ProcessTBench synthetic dataset. The arguments column represents the arguments of the action executions. Actions}
    \label{tab-eventlog}
\end{table}

\vspace{-1.7 cm}
\section{Use-cases}
\vspace{-0.2 cm}
In this section, we provide some of the potential applications of the ProcessTBench synthetic dataset.
\vspace{-0.25 cm}
\paragraph{Evaluating Plan Generation by LLMs:}
The synthetic dataset offers a comprehensive platform to assess how efficiently and accurately LLMs can generate action plans for complex tasks. It enables the exploration of how these models interpret queries, utilize available tools, and sequence actions to solve problems.
\vspace{-0.25 cm}
\paragraph{Evaluating Paraphrase Handling in Plan Generation by LLMs:}
ProcessTBench provides a unique opportunity to assess how well LLMs handle paraphrased or multi-language queries. The synthetic dataset features queries in multiple languages and paraphrased forms, thereby allowing the evaluation of LLMs' versatility and adaptability in different linguistic contexts.
\vspace{-0.25 cm}
\paragraph{Using Process Mining Algorithms for Plan Generation:}
The synthetic dataset can be employed to study the application of process mining techniques in generating and analyzing LLM plans. This involves examining common patterns, deviations, and anomalies in the plans, contributing to a deeper understanding of process behaviors and potential areas for improvement.
\vspace{-0.25 cm}
\paragraph{Plan Generation Variability and Reliability by LLMs:}
The ProcessTBench synthetic dataset allows for an evaluation of the diversity and reliability of plans generated by LLMs. This is crucial for assessing their potential in automating complex tasks, ensuring that the generated plans are not only varied but also consistently accurate and reliable.

\vspace{-0.35 cm}
\section{Conclusion}

ProcessTBench offers a platform for evaluating LLMs in complex plan generation scenarios. By incorporating multilingual query paraphrasing and generating multiple plan variants, this synthetic dataset allows for a more nuanced analysis of LLM behavior, including performance in multi-prompt situations, process discovery, and conformance checking. Future work will further focus on expanding the synthetic dataset with additional queries, languages, plan generation techniques and more sophisticated LLM frameworks.
\vspace{-0.25 cm}
\bibliographystyle{splncs04}
\bibliography{references}   

\begin{thebibliography}{1}
\providecommand{\url}[1]{\texttt{#1}}
\providecommand{\urlprefix}{URL }
\providecommand{\doi}[1]{https://doi.org/#1}

\bibitem{chang_survey_2024}
Chang, Y., Wang, X., Wang, J., Wu, Y., Yang, L., Zhu, K., Chen, H., Yi, X., Wang, C., Wang, Y., Ye, W., Zhang, Y., Chang, Y., Yu, P.S., Yang, Q., Xie, X.: A {Survey} on {Evaluation} of {Large} {Language} {Models}. ACM Trans. Intell. Syst. Technol.  \textbf{15}(3),  39:1--39:45 (Mar 2024). \doi{10.1145/3641289}, \url{https://dl.acm.org/doi/10.1145/3641289}

\bibitem{lassen_complexity_2009}
Lassen, K.B., van~der Aalst, W.M.P.: Complexity metrics for {Workflow} nets. Information and Software Technology  \textbf{51}(3),  610--626 (Mar 2009). \doi{10.1016/j.infsof.2008.08.005}, \url{https://www.sciencedirect.com/science/article/pii/S0950584908001092}

\bibitem{shen_taskbench_2024}
Shen, Y., Song, K., Tan, X., Zhang, W., Ren, K., Yuan, S., Lu, W., Li, D., Zhuang, Y.: {TaskBench}: {Benchmarking} {Large} {Language} {Models} for {Task} {Automation} (Mar 2024), \url{https://openreview.net/forum?id=ZUbraGNpAq}

\bibitem{van_der_aalst_process_2016}
Van Der~Aalst, W.: Process {Mining}. Springer Berlin Heidelberg, Berlin, Heidelberg (2016). \doi{10.1007/978-3-662-49851-4}, \url{http://link.springer.com/10.1007/978-3-662-49851-4}

\bibitem{xu_tool_2023}
Xu, Q., Hong, F., Li, B., Hu, C., Chen, Z., Zhang, J.: On the {Tool} {Manipulation} {Capability} of {Open}-sourced {Large} {Language} {Models} (Nov 2023), \url{https://openreview.net/forum?id=d5ogyvdl1X}

\bibitem{yao_react_2022}
Yao, S., Zhao, J., Yu, D., Du, N., Shafran, I., Narasimhan, K.R., Cao, Y.: {ReAct}: {Synergizing} {Reasoning} and {Acting} in {Language} {Models} (Sep 2022), \url{https://openreview.net/forum?id=WE_vluYUL-X}

\end{thebibliography}
\end{document}